\documentclass[11pt]{article}
\usepackage{geometry} 
\usepackage[backend=biber,style=authoryear,dateabbrev=false, citestyle=authoryear, maxbibnames=99, dashed=false]{biblatex}
\usepackage{hyperref}
\usepackage{float}
\usepackage{times}
\usepackage{latexsym}
\usepackage{array}
\usepackage{coling2020}

\DeclareUnicodeCharacter{0301}{\'{\i}}

\usepackage{url}
\usepackage{microtype}

\colingfinalcopy 

\defbibheading{bibliography}[\refname]{\section*{#1}}

\usepackage{xcolor}

\title{PUM at SemEval-2020 Task 12: Aggregation of Transformer-based models’ features for offensive language recognition}
\author{Piotr Janiszewski, Mateusz Skiba, Urszula Walińska \\
  Faculty of Computing and Telecommunications\\
  Poznan University of Technology \\
  Poznan, Poland \\
  {\tt \{1piotr.janiszewski,xmateusz.skiba,urszula.walinska96\}@gmail.com} \\}

\date{\today}

\begin{document}

\maketitle

\begin{abstract}
In this paper, we describe the PUM team's entry to the SemEval-2020 Task 12. Creating our solution involved leveraging two well-known pretrained models used in natural language processing: BERT and XLNet, which achieve state-of-the-art results in multiple NLP tasks. The models were fine-tuned for each subtask separately and features taken from their hidden layers were combined and fed into a fully connected neural network. The model using aggregated Transformer features can serve as a powerful tool for offensive language identification problem. Our team was ranked 7th out of 40 in Sub-task C - Offense target identification with 64.727\% macro F1-score and 64th out of 85 in Sub-task A - Offensive language identification (89.726\% F1-score).
\end{abstract}


\blfootnote{
    \hspace{-0.65cm}  
    This work is licensed under a Creative Commons 
    Attribution 4.0 International Licence.
    Licence details:
    \url{http://creativecommons.org/licenses/by/4.0/}.
}

\section{Introduction}
Nowadays, abusive content spreading in social media poses a serious problem. The offensive language is pervasive and due to the users' anonymity on the Internet, people feel unreachable when it comes to the punishment. As a result, many people may suffer from being insulted. Multiple social media platforms and companies noticed the emerging problem and have already investigated possible solutions. One of them is to use computational methods to detect hate speech in various types of utterances posted on the web, what has already drawn considerable attention as evidenced in recent publications \parencite{hate1, hate2, hate3}. Nevertheless, the ability of identifying offense automatically is the first step which has to be undertaken. 

In this paper we tackle the problem of Identifying and Categorizing Offensive Language in Social Media. It is the 12-th task of SemEval-2020 competition \parencite{zampieri-etal-2020-semeval}. This is also the second edition of OffensEval, which started in 2019 \parencite{zampieri2019semeval}.

The problem is to detect and assign offensive language occurrences to various pre-defined categories using Natural Language Processing methods. A dedicated datasets consisting of tweets crawled from Twitter were created specifically for this task. Organizers of the competition provided the participants also with the dataset from previous year's edition -- OLID \parencite{zampieri2019predicting}, which we used during the development of our solution. 

Each example is appropriately labelled according to a specific subtask:
\begin{itemize}
    \item \textbf{Sub-task A} -- Identifying if an utterance is offensive (OFF) or not (NOT),
    \item \textbf{Sub-task B} -- Categorizing if a tweet is a targeted (TIN) or untargeted (UNT) insult or threat,
    \item \textbf{Sub-task C} -- Identifying if the offensiveness target is individual (IND) or group (GRP) or other (OTH) e.g. a situation or an event.
\end{itemize}

OffensEval 2020 features a multilingual dataset with five languages. The languages included are: Arabic, Danish, English, Greek and Turkish \parencite{arabic, danish, english, greek, turkish}. From these 5 languages the team members know only English, thus we decided to limit the consideration to this language.

Since training dataset from this year consists of examples with confidence measures produced by unsupervised learning methods instead of labels, we decided to map the scores to the labels. In order to achieve that we simply apply the threshold of 0.4 to the \emph{AVG\_CONF} value in the training phase.

The number of tweets in training and test datasets for subtasks in which we participated are presented in Table~\ref{table:datasetsCardinality}.

Moreover, we had to tackle also a significant problem in both subtasks namely, class imbalance. For example taking into account provided OLID dataset, as for subtask A, it contained 8840 examples labelled as not offensive, whereas only 4400 examples were offensive (proportion 2:1). As for subtask C, the imbalance was even more significant, 2407 examples were labelled as individual offensiveness target, 1074 as group target and only 395 as other (proportion 6:3:1).

\begin{table}
\centering
\caption{The sizes of training and testing datasets in individual tasks}

\vspace{1em}
\begin{tabular}
{ |>{\raggedright\arraybackslash} p{2cm} |
   >{\raggedleft\arraybackslash} p{2cm} |
   >{\raggedleft\arraybackslash} p{2cm} |
   >{\raggedleft\arraybackslash} p{2cm} |
  }
 \hline
 \textbf{Subtask} & \textbf{Training dataset} & \textbf{Testing dataset} & \textbf{OLID dataset}\\
 \hline
 Subtask A & 9075418 & 3887 & 13240\\
 Subtask C & 188973 & 850 & 3876\\
 \hline
\end{tabular}

\label{table:datasetsCardinality}
\end{table}

\section{Methodology}
For development of our solution we used OLID dataset. 90\% of the dataset served for training models and the rest was used as validation set for choosing the best hyperparameters. As for the evaluation of our models' performance we used test dataset from the last year's edition of OffensEval (with 860 examples).
\subsection{Preprocessing}
Our dataset preprocessing started with removing all html tags from the tweets, as we assumed that they did not contain any significant semantic information. We also converted all contractions to full forms, in order not to force our tokenizer to create separate tokens for words like e.g “isn’t” and “is not”. It avoided creating different embeddings for words that have the same meaning. The tweets contained a lot of hashtags too, which were not any specific english dictionary words. In order to make it easier for our models to understand hashtags, we split them with spaces into separate words and removed hashes. For example \#dinnertime would be replaced by two tokens: \emph{dinner}, \emph{time}. Since the tokenizer we used did not recognize emojis, we replaced them with a corresponding text (usually one or two words) that expressed their meaning (e.g. “:)” was replaced by “smile”). We also removed all redundant whitespaces and all accented characters that could mislead our model when creating word embeddings (e.g. \emph{café} was converted to \emph{cafe}). In addition, we also performed filtering sequences of “@user” tokens that were longer than 3, as they were only placeholders for real user nicknames and they did not contain any vital information. 

\subsection{Overcoming dataset imbalance}
As the given provided dataset was highly imbalanced, it was reasonable to measure performance of models using F1-score. This metric takes into account the robustness of a model against data imbalance as it is the harmonic mean of precision and recall. 

Consequently, we decided to find an appropriate method which would be suited for our language model and would help us to overcome the problem. Article \parencite{cost-sensitive} shows that cost-sensitive methods allow the model to achieve better results on F1-score than while using sampling methods. We used the method of levelling instances importance proposed by \parencite{imbalance}. Each instance was assigned with a weight $w = \frac{1}{N * C_i}$, where N is the total number of classes and $C_i$ is the number of instances of the class which the i-th example belongs to. Due to such a solution classes are equally important for the model irrespective of their size.

\subsection{BERT} \label{bert-label}

BERT (Bidirectional Encoder Representation from Transformer) is a model developed by \parencite{bert}. At the moment of publication, it presented state-of-the-art results in wide variety of Natural Language Processing tasks and it is still widely used and developed in numerous solutions. Therefore, we decided to apply it to the OffensEval 2020 problem.

The main feature of BERT is using an attention model called Transformer \parencite{transformer}, which has an ability to learn contextual relations between words and sentences. The Transformer consists of two elements: an encoder which reads text input and embedds it into vectors, and a decoder that outputs a prediction for a task. We used BERT to create a language model, and thus employ only its former part. The most vital aspect of the Transformer is the fact, that it loads an entire sequence of word tokens at once, instead of reading them only from left to right or from right to left. The mechanism enables discovering a context of a word based on its surrounding and this is also the reason why one calls it bidirectional. 

We used a pretrained BERT-Large, Uncased with 24 layers of size 1024 and 16 self-attention heads from Tensorflow Hub \parencite{tensorflow2015-whitepaper}. On the top of it, we put a multilayer perceptron with 3 fully connected layers, first 2 with ReLU activation and sigmoid (for subtask A) or softmax (for subtask C) activation for output. The output score was interpreted as a probability that an example was an offensive online utterance. In case of task A, the output of the sigmoid was interpreted as a probability of an example belonging to the class OFF. In case of task C, the output was a probability distribution over all three classes: IND, GRP, OTH. In order to train the obtained neural network, we used Adam optimizer \parencite{adam} with the default parameter settings. We trained the model for 3 epochs. Following \parencite{fine-tuning}, we did not freeze BERT layers, enabling them to fine-tune to the considered problem instead.

\subsection{XLNet}

XLNet \parencite{xlnet} is a generalized autoregressive model where a next token is dependent on all previous tokens.
XLNet, unlike BERT, can train a model to incorporate bidirectional context, while avoiding the \emph{[MASK]} token and parallel independent predictions by introducing a variant of language modeling called \emph{permutation language modeling}. 
This model is trained to predict a single token given preceding context, but instead of predicting the tokens in sequential order, they are predicted in a random order.
Moreover, XLNet improves upon BERT by using the Transformer XL as its base architecture. The Transformer XL showed state-of-the-art performance in language modeling \parencite{transformer-xl}.

Transformer XL introduces the notion of relative positional embeddings. Instead of having an embedding which represents the absolute position of a word, the Transformer XL uses an embedding to encode the relative distance between words. This embedding is used while computing the attention score between any two words. To sum up, relative positional embedding enables the model to learn how to compute the attention score for words that are \emph{n} words before and after the current word.

\par We used a pretrained \emph{XLNet-base} model with 12 hidden layers of 768 units each and 12 attention heads.
On the top of it, we put 2 fully connected layers with hyperbolic tangent activation for output. We used the same training protocol as described in \ref{bert-label}.

\subsection{Aggregation of models' features}
In general, the purpose of this operation is to combine individual models to obtain a more flexible and powerful classifier. This may resemble one of a category of ensemble methods intended for the models of high bias. These models are not flexible enough to fit into the data and by using an appropriate ensemble method we can obtain a stronger learner.
However, the method of feature aggregation is more sophisticated as it uses neural network to aggregate features derived from different models before making any prediction. The process of running this method can be divided into following steps:

\begin{enumerate}
    \item We create individual models and train them separately for the classification task of offensive language identification as it is described in the previous sections. In our case it is a single BERT model, and a single XLNet model.
    \item For each model, we cut out the dense part responsible for the classification. We make prediction on the whole dataset using so obtained models and proceed with the features from their last hidden layer. In our case, features derived from both models BERT and XLNet are of size 768 (for each example) and they will be frozen from now on.
    \item We create a composite model and its task will be to combine derived features in a relevant way. This model consists of separate inputs for the individual models' frozen features, then a concatenation layer from \textit{Keras} is added to aggregate all of the obtained features into a single feature vector. 
    Lastly, we attach also the dense part for classification. Namely, 3 dense layers with, respectively, 256, 128 and 1 or 3 neurons (1 for binary classification problem -- sub-task A and 3 for problem with 3 classes -- sub-task C). First 2 dense layers have a \textit{ReLU} activation function and the last one has a \textit{sigmoid} or \textit{softmax} activation function, for sub-task A and sub-task C respectively. Between dense layers we add dropout with the rate of 0.1 to prevent the model from overfitting.
    \item We train so obtained network using Adam optimizer \parencite{adam} with default parameters. We use also appropriately weighted binary cross-entropy loss (for binary classification, sub-task A, or weighted categorical cross-entropy loss for sub-task C) to handle the problem of class imbalance. We stop the training earlier based on the best value of F1 metric on validation set.
\end{enumerate}

We hope that since models used in our method differ, their prediction (and features derived from them) will also differ. The models might have paid attention to different aspect of the data throughout the training process. Combining the knowledge from them might enrich the prediction and improve the final result.


\section{Experiments}
In order to check the performance of designed model and compare it to the baselines, we conducted the experiments presented in the section below. We used the test set from last year's competition edition for subtask A, as it also consisted of tweets and concerned recognizing if an utterance was offensive or not.

\subsection{Ensemble methods and feature aggregation evaluation}

In this section, the comparison of individual models, basic ensemble methods and aggregation of models' features is presented. The evaluation is based on the results obtained by each method using a test set according to F1-score. The results are presented in the Table ~\ref{table:ensembleEvaluation}. 

We decided to base our experiment on Transformer models' (BERT and XLNet, both in base version) individual performance and their combination in the form of an ensemble and aggregation of their features. As for ensemble methods we used Soft Majority Voting (method widely used in many applications, e.g. \parencite{inproceedings}) and Logistic Regression (also commonly used in many cases \parencite{article}), as meta-model which took predictions of the classifiers as an input and outputs the probability of the considered example belonging to the class OFF.

\begin{table}[H]
\centering
\caption{Evaluation of individual models, ensemble methods and feature aggregation}
\vspace{1em}
\begin{tabular}{ | p{5cm} | p{1.5cm} | }
\hline
 \textbf{Method} & \textbf{F1-score} \\
 \hline
 BERT (base) & 0.62 \\
 XLNet (base) & 0.69 \\
 \hline
 Soft Majority Voting & 0.68 \\
 Logistic Regression & 0.70 \\
 \textbf{Aggregation of features} & \textbf{0.73} \\
 \hline
\end{tabular}
\label{table:ensembleEvaluation}
\end{table}

We observe that aggregation of models' features was the best method among considered ones. This kind of models' combination was of highest flexibility and therefore enabled us to obtain the best result, which was F1-score equal to 0.73. This method was able to perform well even when the combined models yielded results of different quality (as it is in our case, BERT is worse than XLNet).

The two ensemble methods did not perform as well as the aforementioned aggregation of features. In case of Soft Majority Voting the result was even worse than XLNet individually, which may have been caused by the imbalance of models' results. The solution could be, for instance, assigning BERT a lower weight, so that XLNet's prediction is more important.

Since training Logistic Regression model to combine individual models' predictions provides a bit more freedom and flexibility, the results were slightly better than the ones obtained using Soft Majority Voting.


\subsection{Final evaluation}
In order to meaningfully evaluate the proposed method, we performed a comparison with a set of baseline models. The obtained results are presented in \autoref{table:evaluationIndividualModels}.

\begin{table}[H]
\centering
\caption{Final evaluation}
\vspace{1em}
\begin{tabular}{ | p{5cm} | p{1.5cm} | }
\hline
 \textbf{Model} & \textbf{F1-score} \\
 \hline
 tf-idf + Naive Bayes & 0.56 \\
 tf-idf + SVC & 0.60 \\
 fastText \parencite{joulin2016bag} & 0.61 \\
 GloVe + LSTM & 0.64 \\
 BERT (large) & 0.67 \\
 XLNet (base) & 0.69 \\
 \textbf{BERT and XLNet aggregation of features \textbf{(ours)}} & \textbf{0.73} \\
 \hline
\end{tabular}
\label{table:evaluationIndividualModels}
\end{table}

The outcome of the experiment confirm the observations of other researchers that Transformer models outperform other older Natural Language Processing methods, in particular in the sentiment analysis task. What is more, the aggregation of features obtained by these models may yield even better results.

\section{Competition results}
We took part in sub-tasks A and C in the OffensEval 2 competition. Taking into consideration the results we obtained on the test set (from dataset OLID), the final solution we designed (for both sub-tasks) was based on the model performing the aggregation of features derived from BERT (large) and XLNet (base). Model was trained on dataset OLID and training dataset from OffensEval 2020. Our results including obtained F1-scores and places in the official rank are presented in the Table ~\ref{table:resultsSubtasks}.

\begin{table}[H]
\centering
\caption{Our results in individual subtasks}
\vspace{1em}
\begin{tabular}{ | p{3cm} | p{1.5cm} | p{1.5cm} | }
\hline
 \textbf{Subtask} & \textbf{F1-score} & \textbf{place} \\
 \hline
 Subtask A & 0.897 & 64th \\
 Subtask C & 0.647 & 7th \\
 \hline
\end{tabular}
\label{table:resultsSubtasks}
\end{table}

\section{Conclusions}


In this paper, we considered the problem of deciding whether a short text utterance in English (e.g., a tweet) is offensive or not.
Moreover, for offensive utterances, we considered more detailed classification to decide on the target of the aggression, whether it is an individual, a group or other entity.
We used the dataset provided within SemEval-2020 Task 12 and considered numerous machine learning models.
In particular, we considered BERT and XLNet and showed that their aggregation based on neural networks outperforms any of the models separately, as well as their simpler aggregations based on voting or logistic regression. As a conclusion we observed, that ensemble learning achieved promising results because this method reduces generalization error and is more capable to learn, comparing to single models.
We also performed a comparison with more traditional approaches to NLP, and showed that the obtained aggregate model consistently outperforms them by a wide margin.

The proposed solution ranked 7th in subtask C with the F1 score of $0.647$ and 64th in subtask A with the F1 socre of $0.897$.
We believe that introducing additional models to the aggregation would allow us to achieve even better results.

\section*{Acknowledgements}
We are grateful to Prof. Krzysztof Krawiec of Poznan University of Technology for his invaluable help in the early development stages. We also would like to thank Jedrzej Potoniec of Poznan University of Technology for his help with editing this paper.

\printbibliography
\end{document}